\documentclass{article}
\usepackage{spconf,amsmath,epsfig}
\usepackage{amsfonts}
\usepackage{algorithm, algorithmic}

\usepackage[title]{appendix}
\usepackage{color}
\usepackage{multirow}
\usepackage{amsmath}
\usepackage{breqn}
\usepackage{amssymb}

\let\OLDthebibliography\thebibliography
\renewcommand\thebibliography[1]{
  \OLDthebibliography{#1}
  \setlength{\parskip}{0pt}
  \setlength{\itemsep}{0pt plus 0.3ex}
}

\DeclareMathOperator*{\argmax}{argmax}
\begin{document}\sloppy

\def\x{{\mathbf x}}
\def\L{{\cal L}}

\title{Utility-Maximizing Bidding Strategy for Data Consumers in Auction-based Federated Learning}
%
\name{Xiaoli Tang, Han Yu}
\address{School of Computer Science and Engineering, Nanyang Technological University, Singapore\\
\{xiaoli001, han.yu\}@ntu.edu.sg
}

\maketitle

\begin{abstract}
Auction-based Federated Learning (AFL) has attracted extensive research interest due to its ability to motivate data owners to join FL through economic means. Existing works assume that only one data consumer and multiple data owners exist in an AFL marketplace (i.e., a monopoly market). Therefore, data owners bid to join the data consumer for FL. However, this assumption is not realistic in practical AFL marketplaces in which multiple data consumers can compete to attract data owners to join their respective FL tasks. In this paper, we bridge this gap by proposing a first-of-its-kind utility-maximizing bidding strategy for data consumers in federated learning (Fed-Bidder). It enables multiple FL data consumers to compete for data owners via AFL effectively and efficiently by providing with utility estimation capabilities which can accommodate diverse forms of winning functions, each reflecting different market dynamics. Extensive experiments based on six commonly adopted benchmark datasets show that Fed-Bidder is significantly more advantageous compared to four state-of-the-art approaches.
\end{abstract}

\section{Introduction}
\label{sec:introduction}
Due to user privacy and data confidentiality requirements, Federated Learning (FL) has attracted significant research interest \cite{yang2019federated,Liu-et-al:2020FedVision,Liu-et-al:2022IAAI}. 
Auction-based FL (AFL) has become an important type of FL in recent years. In general, data consumers (DCs) initiate FL tasks for data owners (DOs) to join. Existing works can be divided into three categories:
1) the \textit{supply side problem} studies how a DO determines the amount of resources to commit to FL and sets the reserve price to maximize its profit; 2) the \textit{auctioneer problem} studies the optimal DC-DO matching and pricing in order to achieve desirable operational objectives (e.g., social welfare maximization, social cost minimization) for the AFL ecosystem, and 3) the \textit{demand side problem} studies how DCs select and bid for DOs to maximize specific key performance indicators (e.g., accuracy improvement) within budget constraints.

Existing AFL approaches mostly belong to the first two categories. State-of-the-art works belonging to the third category \cite{zhang2021incentive,zhang2022auction,zhang2022online}  are based on a common limiting assumption: \textit{there is only one DC and multiple DOs in an AFL marketplace (i.e., a monopoly market)} \cite{Shi-et-al:2023FAFL}.
In practice, especially open collaborative AFL marketplaces in which multiple DCs can initiate FL tasks simultaneously and compete with each other to enlist DOs \cite{Lyu-et-al:2020TPDS}, this assumption is not realistic. 

To relax this limiting assumption, 
we define the demand side problem in a competitive AFL market as designing the optimal bidding strategy (i.e., the optimal bidding function) to help each DC generate a customized bid value for each FL DO. 
We propose a utility-maximizing bidding strategy for DCs in AFL (Fed-Bidder) to obtain the optimal bidding functions for DCs
It takes into account not only their limited budgets and the suitability of DOs, but also prior auction-related knowledge (e.g., the distribution of the DOs, the probability of the DC winning the ongoing auction). 
We show that both the estimation of DOs' utility and the appropriate winning function play key roles in determining the optimal bidding function under Fed-Bidder. 
To effectively solve the optimal bidding function, we design a utility estimation algorithm and introduce two representative winning functions \cite{zhang2014optimal} into the AFL ecosystem. 

To the best of our knowledge, Fed-Bidder is the first bidding approach designed to support multiple DCs to compete for a same pool of candidate DOs simultaneously, thereby facilitating more realistic open AFL marketplaces to emerge.
Extensive experiments based on six commonly adopted benchmark datasets show that Fed-Bidder is significantly more advantageous compared to four state-of-the-art approaches, outperforming the best baseline by 12.11\%, 21.87\% and 1.57\% on average in terms of the total amount of data attracted, unit price per 1,000 data samples and FL model accuracy, respectively. 



\section{Related Work}
\label{sec:related_work}
Existing auction-based incentive mechanisms in FL could be roughly grouped into two main categories, 1) those designed for the auctioneer to maximize social welfare or minimize social cost, and 2) those designed to help the DC select high-quality DOs and determine the rewards for them. 

Methods falling into the first category generally use the combinatorial auction mechanism to cope with issues with more than one DC competing for multiple different but related resources owned by multiple DOs. The typical method of this category is \cite{yang2020task}. In this work, the combinatorial auction-based market model is adopted to distribute tasks from DCs to DOs. In this model, various consumers submit diverse requirements and bid prices for providers in sequence. After receiving the requests from the consumers, the winner determination algorithm is adopted to determine the winning consumers and their corresponding payments. The optimal consumer-provider matching relationship is transformed into a multidimensional grouping knapsack problem, and then solved with dynamic programming.

Methods in the second category \cite{zhang2021incentive,zhang2022auction,zhang2022online, le2020auction,le2021incentive,roy2021distributed,deng2021fair,ying2020double,zeng2020fmore,jiao2020toward} utilize the reverse auction mechanism, combined with various techniques and mechanisms, such as reputation, blockchain, graph neural networks as well as deep reinforcement learning, to cope with issues where one specific DC wants to achieve its desirable objectives. Specifically, Zhang et al. \cite{zhang2021incentive} incorporated reputation and blockchain into reverse auction and propose the RRAFL approach. In this work, the reputation, which is determined by the DC based on DOs’ reliability and data quality track records stored in blockchains, is used to help the target DC determine the winning DOs.

The first category is based on an FL market setting with multiple DCs and DOs. However, methods in this category focus on the objectives of the auctioneer, ignoring how DCs shall bid. This paper is for DCs, closer to the second category of methods. Existing methods falling into the second category are based on the common assumption of a monopoly FL market in which only one DC exists, which is not realistic in FL marketplaces where multiple DCs can exist simultaneously and compete for the same pool of DOs. The proposed Fed-Bidder approach bridges this gap in the literature. 


\section{Preliminaries}
\label{sec:preliminaries}

\begin{figure}[t!]
\centering
\includegraphics[width=0.8\linewidth]{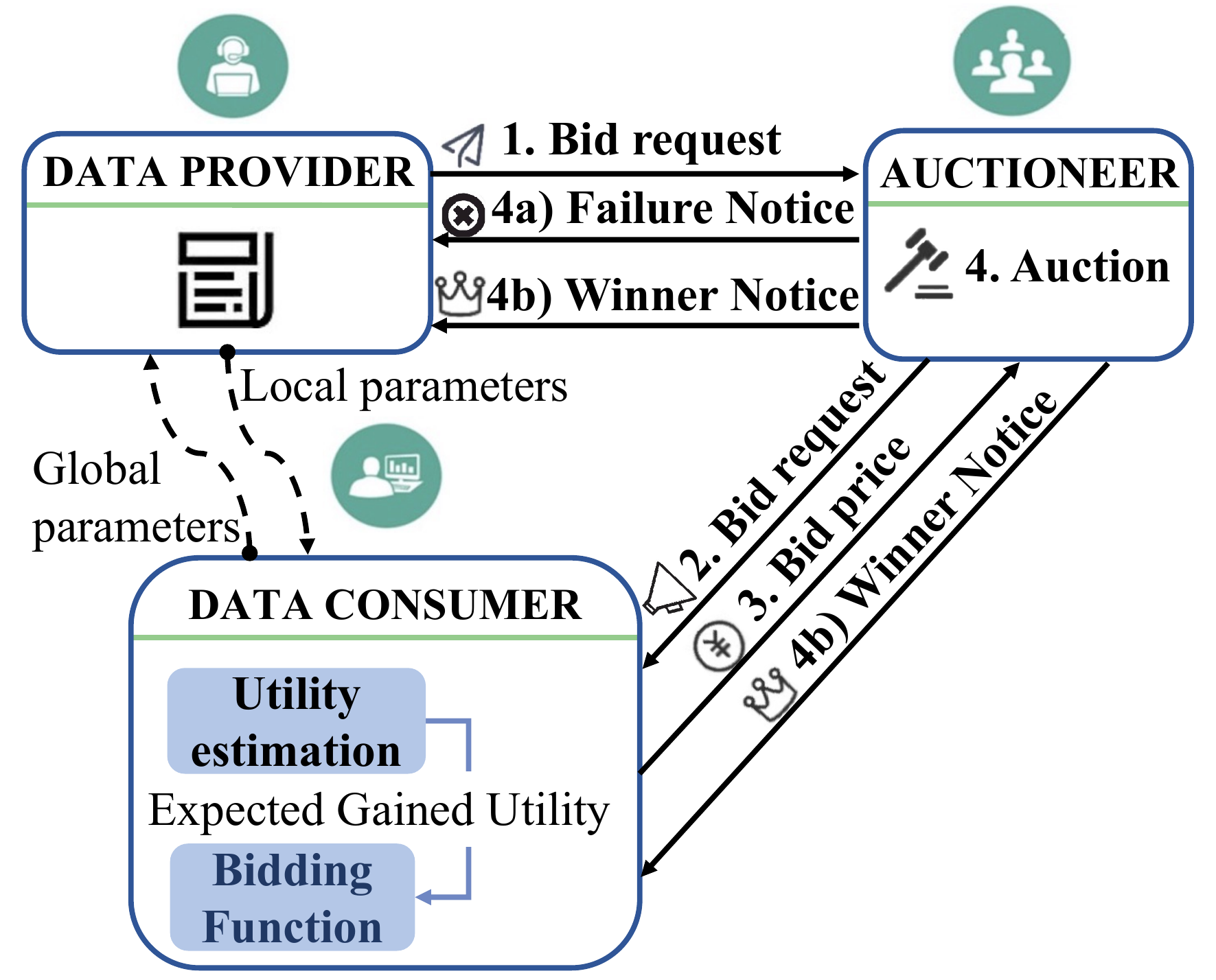}
\caption{
The auction workflow triggered by an interested DO.}
\label{fig:trading_process}
\end{figure}

\textbf{System Model}:
\label{sec:system_model}
The FL ecosystem of focus includes: 1) \textit{DOs} (i.e., the supply side), 2) \textit{DCs} (i.e., the demand side), and an \textit{FL auctioneer}.
Whenever a qualified DO has eligible data resources, it can indicate its interest to join the given DC's auction process. Each DC then bids for the interested DOs. Following \cite{zhang2022online}, we assume that the resources of each qualified DO become gradually available before or during the processing of tasks from various DCs. Fig. \ref{fig:trading_process} shows the workflow of an auction.
The auctioneer can hold such an auction process for each bundle of local data resources owned by a DO, 
and broadcast them to each DC. When a data consumer has recruited enough DOs or has exhausted its budget, it exits the auction and initiates FL model training with the DOs it has recruited.
Finally, each DC pays the DOs it has recruited.

\noindent \textbf{Key Assumptions}:
\label{sec:notations}
Suppose there are $N$ qualified bid requests in an FL ecosystem, for which all interested DCs can bid. For simplicity, we suppose each provider generates only one bid request. 
Each bid request triggered by the DO is denoted as a high-dimensional feature vector $\boldsymbol{q}_i$, where the entries comprise features related to the DO (e.g., the identity of the DO and the data quantity). Moreover, we suppose the bid request $\boldsymbol{q}_i$ follows identical independent distribution (i.i.d.) from a prior distribution $p_{\mathcal{Q}(\boldsymbol{q})}$. After receiving the bid request $\boldsymbol{q}_i$ transmitted by the auctioneer, each DC $j$ can estimate the potential utility $s_j(\boldsymbol{q}_i)$ if it wins the auction. 
Similar to $\boldsymbol{q}_i \sim p_{\mathcal{Q}(\boldsymbol{q})}$, 
$s_j(\boldsymbol{q}_i) \sim p_{\mathcal{S}}(s_{j,i})$ 
Based on the estimated utility, the DC $j$ calculates its bid price $b_{j,i}$ for the auction $i$ according to the bidding strategy with a bidding function $b_j(s(\boldsymbol{q}_i))$. The probability of winning the auction is determined by the winning function $W_j(b(s(\boldsymbol{q}_i)))$\footnote{For composite functions (i.e., $b_j(s_j(\boldsymbol{q}_i))$ and $W_j(b_j(s_j(\boldsymbol{q}_i)))$), we only keep the subscript $j$ for the outermost function (i.e., $b_j(s(\boldsymbol{q}_i))$ and $W_j(b(s(\boldsymbol{q}_i)))$) when there is no risk of confusion}. 

\section{The Proposed Approach}
\label{sec:model}

\subsection{Decision Support for Data Consumers}
In this paper, we focus on the demand side problem in auction-based FL. 
Thus, the decision support provided by Fed-Bidder mainly revolves around bid calculation for each bid request, the solution to which is referred to as the bidding strategy.
In particular, we aim to design a bidding function
$b^{O}_j()$ to help a DC $j$ bid for DOs with appropriate bid prices so as to optimise its utility within the budget limit in a competitive setting. The problem is formulated as:  
\begin{equation}
\label{eq:original_text_goal}
\begin{aligned}
b^{O}_j() \triangleq \argmax_{b_j()} TotalUtility_j,\\
\quad s.t. \quad TotalCost_j \leqslant budget_j, 
\end{aligned}
\end{equation}
where \textit{TotalUtility$_j$} and \textit{TotalCost$_j$} refer to the overall expected utility and overall cost of the DC $j$, respectively, under the given bidding strategy. 

Given $N$ eligible qualified bid requests and the budget $B_j$ of DC $j$, the objective described in Eq. \eqref{eq:original_text_goal} can be re-expressed as an expectation maximization formulation:
\begin{equation}
\label{eq:text_goal_1}
\begin{aligned}
b^{O}_j() = \argmax_{b_j()}N\int_{q_i} T_j(\boldsymbol{q}_i) p_{\mathcal{Q}(\boldsymbol{q})}d\boldsymbol{q}_i,\\
\quad s.t. \quad N\int_{q_i} C_j(\boldsymbol{q}_i)p_{\mathcal{Q}(\boldsymbol{q})}d\boldsymbol{q}_i \leqslant B_j.
 \end{aligned}
\end{equation}
We set the bid price $b_j(s(\boldsymbol{q}_i))$ as the upper boundary of the winning cost following \cite{zhang2014optimal}. Therefore, in Eq. \eqref{eq:text_goal_1}, $T_j(\boldsymbol{q}_i)$, which represents the expected utility the target DC $j$ can gain from the bid request $\boldsymbol{q}_i$ with the bid price $b_j(s(\boldsymbol{q}_i))$, is expressed as:
$T_j(\boldsymbol{q}_i)\triangleq(s_j(\boldsymbol{q}_i)-b_j(s(\boldsymbol{q}_i)))W_j(b(s(\boldsymbol{q}_i)))$.
$s_j(\boldsymbol{q}_i)-b_j(s(\boldsymbol{q}_i))$ denotes the utility gained by DC $j$. $C_j(\boldsymbol{q}_i)$ from Eq. \eqref{eq:text_goal_1} can be expressed as 
 $C_j(\boldsymbol{q}_i)\triangleq b_j(s(\boldsymbol{q}_i))W_j(b(s(\boldsymbol{q}_i)))$.
Thus, Eq. \eqref{eq:text_goal_1} becomes
\begin{equation}
\label{eq:original_goal}
\begin{aligned}
 & b^{O}_j() = \argmax_{b_j()}N\int_{q_i} (s_j(\boldsymbol{q}_i)-b_j(s(\boldsymbol{q}_i)))\\
&\quad \quad \quad \quad \quad \quad \quad \quad \quad \quad \quad \quad W_j(b(s(\boldsymbol{q}_i)))p_{\mathcal{Q}(\boldsymbol{q})}d\boldsymbol{q}_i,\\
& \quad s.t. \quad N\int_{q_i} b_j(s(\boldsymbol{q}_i))W_j(b(s(\boldsymbol{q}_i)))p_{\mathcal{Q}(\boldsymbol{q})}d\boldsymbol{q}_i \leqslant B_j.   
\end{aligned}
\end{equation}
Considering the relationship between $\boldsymbol{q}_i$ and $s_j(\boldsymbol{q}_i)$, we have
$p_{\mathcal{Q}(\boldsymbol{q})}=p_{\mathcal{S}}(s_j(\boldsymbol{q}_i)) \lVert \nabla s_j(\boldsymbol{q}_i) \rVert$. 
Substituting $p_{\mathcal{Q}(\boldsymbol{q})}$
into Eq. \eqref{eq:original_goal} gives, 
\begin{equation}
\label{eq:final_goal}
\begin{aligned}
& b_j^{O}() =\argmax_{b_j()}N\int_{s_{j,i}} (s_{j,i} -  b_{j,i}(s)) \\
& \quad \quad \quad \quad \quad \quad \quad \quad \quad \quad \quad \quad W_{j,i}(b(s))p_{\mathcal{S}}(s_{j,i})ds_{j,i},\\
& \quad s.t. \quad N\int_{s_{j,i}} b_{j,i}(s)W_{j,i}(b(s))p_{\mathcal{S}}(s_{j,i})ds_{j,i} \leqslant B_j.
 \end{aligned}
\end{equation}

To solve the optimisation problem in Eq. \eqref{eq:final_goal}, we first obtain the corresponding Lagrangian:
\begin{equation}
\label{eq:lagrangian_goal}
\begin{aligned}
\mathcal{L}(b_{j,i}(s), \lambda) = \int_{s_{j,i}} (s_{j,i}-b_{j,i}(s)) W_{j,i}(b(s))p_{\mathcal{S}}(s_{j,i})ds_{j,i}\\
\quad \quad -\lambda\int_{s_{j,i}} b_{j,i}(s)W_{j,i}(b(s))p_{\mathcal{S}}(s_{j,i})ds_{j,i} + \frac{\lambda B_j}{N},
 \end{aligned}
\end{equation}
where $\lambda$ is the Lagrangian multiplier. 
This optimization problem can be regarded as a functional extremum problem. Based on the calculus of variations, the necessary condition for finding the extremum of $b_{j,i}(s)$ is that the first order derivative equals to 0, 
, i.e.,
$\frac{\partial \mathcal{L}(b_{j,i}(s), \lambda)}{\partial b_{j,i}(s)}=0$. Then, we get 
$s_{j,i} p_{\mathcal{S}}(s_{j,i}) \frac{\partial W_{j,i}(b(s))}{\partial b_{j,i}(s)} - (\lambda +1) p_{\mathcal{S}}(s_{j,i})
\left[W_{j,i}(b(s)) + b_{j,i}(s)\frac{\partial W_{j,i}(b(s))}{\partial b_{j,i}(s)}\right]=0$, i.e., 
\begin{equation}
\label{eq:b_theta}
\begin{aligned}
(\lambda +1) W_{j,i}(b(s)) = [s_{j,i} - (\lambda +1) b_{j,i}(s)] \frac{\partial W_{j,i}(b(s))}{\partial b_{j,i}(s)}.
 \end{aligned}
\end{equation}
Eq. \eqref{eq:b_theta} shows that the estimated utility $s_{j,i}$ and the winning function $W_{j,i}(b(s))$ are key to finding the optimal bidding function $b_{j,i}(s)$ for DC $j$ to bid request $\boldsymbol{q}_i$. Thus, in the following sections, we describe how to calculate them. 

\subsection{Utility Estimation}
\label{sec:utility_estimation}
Following \cite{zhan2020learning}, we define the utility estimation function of DC $j$ with respect to a given bid request from a DO $i$ as:
\begin{equation}
\label{eq:utility_estimation}
\begin{aligned}
s_j(\boldsymbol{q}_i) \triangleq \ln(1+\theta_j^T\boldsymbol{q}_i) = s_{j,i},
 \end{aligned}
\end{equation}
where $\theta_j$ is a learnable parameter. For clarity, we denote $s_j(\boldsymbol{q}_i)$ as $s_j(\cdot)$ in subsequent derivations.

Over multiple rounds of auctions, DCs in an FL ecosystem can accumulate historical data $H$ which can be used to derive the utility of the current bid request. Such data can be recorded in the form of $(\boldsymbol{q}_m, y_m)$ ($(\boldsymbol{q}_m, y_m) \in H$), where $y_m$ denotes the real utility of bid request $\boldsymbol{q}_m$. Then, we leverage the squared error (SE) loss \cite{ren2017bidding} to train the utility estimation function $s_j(\cdot)$,
$L(s_j(\cdot)) = \frac{1}{2} \sum_{(\boldsymbol{q}_m, y_m)}(y_m-s_j(\boldsymbol{q}_m))^{2}$.
The parameter $\theta_j$ in Eq. \eqref{eq:utility_estimation} can be obtained via gradient descent:
$\theta_j \gets \theta_j - \eta_{\theta_j} \frac{\partial L(s_j(\cdot))}{\partial \theta_j}$,
where $\eta_{\theta}$ is the learning rate and $\frac{\partial L(s_j(\cdot))}{\partial \theta_j}$ is derived as:
$\frac{\partial L(s_j(\cdot))}{\partial \theta_j} =  \sum_{(\boldsymbol{q}_m, y_m)} [ \ln(1+\theta_j^T\boldsymbol{q}_m) -y_m] \frac{\boldsymbol{q}_m}{1+\theta_j^T\boldsymbol{q}_m}$.

\subsection{Winning Functions and Bidding Functions}
The bidding function also depends on the winning function. In \cite{zhang2014optimal,zhang2015statistical}, it has been shown that the winning functions related to real world datasets\footnote{http://data.computational-advertising.org} usually take a concave shape. Thus, we adopt the concave shape to propose two forms of winning functions with different levels of complexity in anticipation of different FL auctioning scenarios, and show how to derive the corresponding optimal bidding functions. 

\noindent \textbf{Simple Concave Functions}:
Following \cite{zhang2014optimal}, the winning rate $W_j(s)$ can be of a simple concave form, which can be expressed as:
\begin{equation}
\label{eq:winning_function_1}
\begin{aligned}
W_{j,i}(b(s)) \triangleq \frac{b_{j,i}(s)}{c_j+b_{j,i}(s)}
 \end{aligned}
\end{equation}
where $c_j$ is a constant set by DC $j$. By differentiating Eq. \eqref{eq:winning_function_1} with respect to the bidding function $b_j(s)$, we have:
$\frac{\partial W_{j,i}(b(s))}{\partial b_{j,i}(s)} = \frac{c_j}{(c_j+b_{j,i}(s))^2}$.
Substituting $W_{j,i}(b(s))$ and $\frac{\partial W_{j,i}(b(s))}{\partial b_{j,i}(s)}$ into Eq. \eqref{eq:b_theta}, we have:
$(c_j+b_{j,i}(s))^2 = c_j^2 + \frac{s_{j,i} c_j}{\lambda +1}$
The ﬁnal optimal bidding function $b_{j,i}^{O}(s)$ can be derived as:
\begin{equation}
\label{eq:final_ob_function_1}
\begin{aligned}
b_{j,i}^{O}(s) = \sqrt{c_j^2 + \frac{s_{j,i} c_j}{\lambda +1}} - c_j.
 \end{aligned}
\end{equation}

\noindent \textbf{Complex Concave Functions}:
In some cases, the winning function can be of a more complex form than in Eq. \eqref{eq:winning_function_1}:
\begin{equation}
\label{eq:winning_function_2}
\begin{aligned}
W_{j,i}(b(s)) \triangleq \frac{b_{j,i}^2(s)}{c_j^2+b_j^2(s)},
 \end{aligned}
\end{equation}
Following the process described above, we obtain the optimal bidding function for Eq.
\eqref{eq:winning_function_2} as:
\begin{equation}
\label{eq:final_ob_function_2}
\begin{aligned}
b_{j,i}^{O}(s) = c_j\left(\frac{s_{j,i}+\sqrt{c_j^2(\lambda +1)^2 + s_{j,i}^2}}{c_j(\lambda+1)}\right)^{\frac{1}{3}} \\
- c_j\left(\frac{c_j(\lambda+1)}{s_{j,i}+\sqrt{c_j^2(\lambda+1)^2+s_{j,i}^2}}\right)^{\frac{1}{3}}.
 \end{aligned}
\end{equation}

\section{Experimental Evaluation}
\label{sec:experiment}

\subsection{Experiment Setup}
\label{sec:experiment_setup}
Six commonly adopted datasets in FL studies are used in this paper: MNIST \cite{lecun1998mnist}, CIFAR-10 and CIFAR-100\footnote{https://www.cs.toronto.edu/~kriz/cifar.html},
Fashion-MNIST (FMNIST) \cite{xiao2017fashion}, EMNIS-digits (EMNIST-D) / letters  (EMNIST-L) \cite{cohen2017emnist}, Kuzushiji-MNIST (KMNIST) \cite{clanuwat2018deep}. 
In each experiment, we create 100 DOs in the FL ecosystem. The training set size of each DO is random, ranging from 1,000 to 10,000 samples. 
Each DC's validation set and test set both contain 2,000 samples. Following \cite{zhang2021incentive}, the FL model to be trained contains an input layer with 784 nodes, a hidden layer with 50 nodes and an output layer with 10 nodes for tasks on MNIST, FMNIST, EMNIST-D and KMNIST. Tasks on EMNIST-L are processed with the aforementioned network structure but with the output layer containing 26 nodes. For tasks on CIFAR,
we use the simplified VGG11 architecture \cite{simonyan2014very}, where the number of convolutional filters and the size of the hidden fully-connected layers are $\{32, 64, 128, 128, 128, 128, 128, 128\}$ and 128, respectively.

To make it possible for DCs to effectively estimate the utility of providers, each DO is named by sequence number. Then, the data of the first half of the providers are blurred  while those of the second half are unchanged. Therefore, DCs could measure the utility of each DO from the perspective of both the quantity and the quality of data, the later of which is reflected by the sequence number. 
Then, we use the auction mechanism proposed by \cite{jin2015auction} to generate historical auctioning data, winning records and auction records. After that, each DC utilizes the DOs it obtained through the current auction to train the FL model. The utility evaluation method in \cite{zhang2021incentive} is adopted to compute the real utility obtained by each DO. Then, based on the real utility of each winning record, each DC can train its own utility estimation function following Eq. \eqref{eq:utility_estimation}. Afterwards, the winning functions shown in Eq. \eqref{eq:winning_function_1} and Eq. \eqref{eq:winning_function_2} as well as the optimal $\lambda$ can be obtained based on the historical winning records as well as the estimated utility. 



\subsection{Comparison Approaches}
For clarity, we refer to the variants of Fed-Bidder in Eq. \eqref{eq:final_ob_function_1} and Eq. \eqref{eq:final_ob_function_2} as \textit{FBs} and \textit{FBc}, respectively. 
We compare FBs and FBc with the following bidding strategies experimentally: 
\begin{enumerate}
    \item \textbf{Constant Bid (Const)} \cite{zhang2014optimal} offers a constant bid for all the bid requests. The bid value by each DC can differ.
    \item \textbf{Randomly Generated Bid (Rand)} \cite{zhang2021incentive,zhang2022online} is commonly used in auction-based FL, which randomly generates a bid from a fixed range of values for each bid request. 
    \item \textbf{Below Max Utility Bid (Bmub)} is adapted from bidding below max eCPC \cite{lee2018estimating} in the field of advertising. For each bid request, the bid price is randomly generated in the range of 0 and its utility.
    \item \textbf{Linear-Form Bid (Lin)} \cite{perlich2012bid} sets the bid price to be linearly related to the estimated utility of the bid request. 
\end{enumerate}

We create six DCs, each adopting one of the aforementioned bidding approach to compete for the same pool of FL DOs. The auction process presented in Section \ref{sec:system_model} is then held for each bid request. 
If there is no more bid request or no budget left, the procedure terminates.

\subsection{Results and Discussion}
\begin{table*}[ht]
\centering
\caption{Bidding performance comparison under different budget settings and datasets. $\#$Total denotes the total number of data samples (the higher, the better), while u.p. denotes the unit price per 1,000 data samples (the lower, the better).} 
\resizebox*{0.8\linewidth}{!}{
\begin{tabular}{|*{16}{c|}}
\hline
\multirow{2}*{Budget} & \multirow{2}*{Method} & \multicolumn{2}{c|}{MNIST} & \multicolumn{2}{c|}{CIFAR-10} &\multicolumn{2}{c|}{FMNIST}& \multicolumn{2}{c|}{EMNIST-D} & \multicolumn{2}{c|}{EMNIST-L} & \multicolumn{2}{c|}{KMNIST} & \multicolumn{2}{c|}{CIFAR-100}\\\cline{3-16}
{} & {} & $\#$Total &  u.p. & $\#$Total &  u.p. & $\#$Total &  u.p. & $\#$Total &  u.p. & $\#$Total &  u.p. & $\#$Total &  u.p. & $\#$Total &  u.p. \\\hline
\multirow{6}*{50}  & Const & 13,878 & 3.24 & 10,222 & 4.30 & 14,898 & 3.22 & 11,047 & 4.44 & 8,728 & 5.15 & 11,021 & 4.45 & 7,143 & 5.71\\
{} & Rand & 6,688 & 7.24 & 9,724 & 4.29 & 16,725 & 2.76 & 9,821 & 4.62 & 9,841 & 4.92 & 10,948 & 4.41 & 7,678 & 6.30\\
{} & Bmub & 14,783 & 3.32 & 13,118 & 3.60 & 17,301 & 2.79 & 11,465 & 4.23 & 11,476 & 3.98 & 8,386 & 5.91 & 8,773 & 5.59\\
{} & Lin & 16,433 & 3.01 & 13673 & 3.65 & 17,627 & 2.81 & 11,628 & 4.27 & 14,289 & 3.02 & 12,423 & 3.84 & 9,423 & 5.25\\\cline{2-16}
{} & FBs & \textbf{17,437} & \textbf{2.79} & \textbf{15,931} & \textbf{3.11} & \textbf{18,447} & \textbf{2.62} & 12,037 & 3.81 & \textbf{14,438} & 2.72 & 13,843 & \textbf{3.27} & \textbf{14,427} & \textbf{3.37}\\
{} & FBc & 17,197 & 2.87 & 14,568 & 3.24 & 18,026 & 2.68 & \textbf{13,359} & \textbf{3.62} & 14,032 & \textbf{2.50} & \textbf{13,979} & 3.47 & 14,187 & 3.43\\\hline
\multirow{6}*{150}  & Const & 35,872 & 4.01 & 17,210 & 6.92 & 35,691 & 4.03 & 20,482 & 6.98 & 33,132 & 4.53 & 25,012 & 5.76 & 15,862 & 9.07\\
{} & Rand & 24,195 & 5.79 & 12,872 & 10.81 & 35,054 & 4.17 & 21,349 & 6.97 & 26,437 & 5.54 & 23,741 & 6.23 & 14,185 & 9.88\\
{} & Bmub & 35,980 & 4.16 & 19,157 & 7.59 & 40,072 & 3.58 & 23,382 & 6.38 & 43,525 & 3.41 & 33,956 & 4.18 & 35,970 & 4.16\\
{} & Lin & 40,378 & 3.71 & 35,378 & 4.15 & 42,314 & 3.48 & 30,967 & 4.57 & 46,358 & 3.07 & 37,358 & 3.77 & 40,368 & 3.71\\\cline{2-16}
{} & FBs & \textbf{50,321} & \textbf{2.98} & \textbf{40,851} & \textbf{3.62} & 53,021 & 2.81 & \textbf{38,459} & \textbf{3.59} & \textbf{50,634} & \textbf{2.88} & \textbf{44,427} & \textbf{3.32}  & \textbf{50,311} & \textbf{2.97}\\
{} & FBc & 46,893 & 3.06 & 39,383 & 3.67 & \textbf{54,052} & \textbf{2.75} & 36,241 & 3.97 & 49,595 & 2.96 & 40,313 & 3.71 & 46,883 & 3.19\\\hline
\multirow{6}*{300}  & Const & 59,072 & 4.88 & 47,215 & 6.12 & 73,772 & 3.9 & 44,675 & 6.38 & 47,768 & 5.34 & 59,754 & 4.99 & 22,062 & 13.05\\
{} & Rand & 59,947 & 4.91 & 54,639 & 5.23 & 40,456 & 7.17 & 43,292 & 6.81 & 56,543 & 4.94 & 55,036 & 5.23 & 20,937 & 14.05\\
{} & Bmub & 65,716 & 4.53 & 60,637 & 4.63 & 77,361 & 3.84 & 50,743 & 5.93 & 58,453 & 4.91 & 67,574 & 4.37 & 35,706 & 8.33\\
{} & Lin & 65,749 & 4.55 & 60,698 & 4.53 & 90,325 & 3.32 & 65,042 & 4.56 & 79,452 & 3.63 & 74,843 & 3.98 & 65,739 & 4.55\\\cline{2-16}
{} & FBs & \textbf{96,821} & \textbf{3.14} & \textbf{66,432} & \textbf{4.41} & \textbf{109,011} & \textbf{2.75} & 65,895 & \textbf{4.47} & 80,454 & 3.61 & \textbf{77,007} & \textbf{3.82} & \textbf{92,811} & \textbf{3.20}\\
{} & FBc & 92,580 & 3.21 & 65,793 & 4.52 & 98,904 & 2.98 & \textbf{66,453} & 4.49 & \textbf{81,537} & \textbf{3.44} & 76,758 & 3.91 & 84,570 & 3.51\\\hline
\end{tabular}
}
\label{tab:bidding_performance}
\end{table*}

\begin{table*}[ht]
\centering
\caption{Accuracy (\%) comparison of the FL models on the 6 datasets under IID and non-IID (denoted as NIID) settings.} 
\resizebox*{0.72\linewidth}{!}{
\begin{tabular}{|*{15}{c|}}
\hline
\multirow{2}*{Method} & \multicolumn{2}{c|}{MNIST} & \multicolumn{2}{c|}{CIFAR-10} &\multicolumn{2}{c|}{FMNIST}& \multicolumn{2}{c|}{EMNIST-D} & \multicolumn{2}{c|}{EMNIST-L} & \multicolumn{2}{c|}{KMNIS}& \multicolumn{2}{c|}{CIFAR-100}\\\cline{2-15}
{} & IID &  NIID & IID &  NIID& IID &  NIID& IID &  NIID& IID &  NIID& IID &  NIID& IID &  NIID\\\hline
Const & 85.52  & 74.35 & 44.58  & 21.63  & 73.18  & 64.19& 81.89  & 77.97 & 74.57  & 66.77 & 71.13  & 62.12  & 40.36  & 18.16\\
Rand & 84.92  & 74.11 & 46.71  & 24.54  & 74.36  & 64.08& 81.28  & 77.28 & 74.89  & 65.54 & 69.54  & 62.63 & 39.58 & 18.62\\
Bmub & 85.63  & 75.04 & 47.44  & 25.97  & 74.52  & 65.15& 83.22  & 78.77 & 74.86  & 66.69 & 72.91  & 67.67 & 40.53 & 17.84\\
Lin & 85.86  & 75.71 & 47.81  & 29.06  & 76.82  & 66.42 & 83.74  & 78.86 & 75.06  & 68.70 & 74.36  & 69.16 & 41.04 & 20.12\\\hline
FBs & \textbf{86.57}  & \textbf{76.76} & \textbf{49.62}  & \textbf{30.18} & \textbf{78.68}  & \textbf{66.80} & \textbf{84.46}  & \textbf{79.78} & \textbf{75.89}  & \textbf{70.46} & \textbf{76.78}  & \textbf{70.97} & 41.83 & \textbf{20.75}\\
FBc & 86.29  & 76.08 & 48.53  & 29.35  & 78.28  & 66.62 & 84.23  & 79.52 & 75.52  & 70.36 & 75.41  & 70.39 & \textbf{41.76} & 20.91\\\hline 
\end{tabular}
}
\label{tab:fl_performance}
\end{table*}

\noindent \textbf{Auction Performance}:
The comparison of the bidding strategies under different budget settings (i.e., low budget of 50, medium budget of 150, and high budget of 300) is shown in Table \ref{tab:bidding_performance}. 
It can be observed that, under all budget settings, the proposed Fed-Bidder achieves the best performance in terms of the total amount of data attracted from DOs (\#Total). Compared with the best performing baseline Lin, Fed-Bidder attracts 12.11\% more total amount of data and decreases the unit price of per 1,000 data samples by 21.87\% when averaged over all six datasets.
Const and Rand achieve the worst performance in this respect as they bid randomly without considering the quantity and quality of the data from DOs. This shows that it is useful to estimate the utility of the bid request before making the bid response. 
Under low budget settings, Fed-Bidder achieves the best performance in terms of both \#Total and the unit price per 1,000 data. In such cases, the available budget for each bid request is relatively low. A good bidding strategy shall be able to lower the bid price for each request. Compared with others, Fed-Bidder can opportunistically focus on bidding more aggressively for the low cost DOs based on the concave shape bidding functions, while jointly considering their local data quantity and quality.
When the budget is increased from low to high, the unit price per 1,000 data achieved by all approaches increase. This can be attributed to the fact that when the budget is high, more budget can be allocated to each bid request. Then, the strategies tend to focus their budgets on bidding for more high cost DOs as they generally bring about higher utility for the FL models. 

\noindent \textbf{FL Model Performance}: 
Table \ref{tab:fl_performance} shows the test accuracy achieved by the FL models trained by different bidding approaches under IID and non-IID settings, respectively. The low accuracy of the FL models on CIFAR-10 can be attributed to the base model adopted.

It can be observed that the FL model based on the proposed Fed-Bidder is the best under all settings and outperforms the best baseline Lin by 1.57\% on average in terms of FL model accuracy. Under Fed-Bidder, the target DC is more competitive compared to other approaches in terms of attracting high quality DOs under given budget constraints. In particular, as analysed above, when the budget is low, Fed-Bidder can guide the target DC to bid for more low cost DOs in order to guarantee enough data.
When the DC has a higher budget, all bidding strategies tend to guide the DC to spend more budget on bid requests with potentially high estimated costs (i.e., those with high quality data and high utility). 
\section{Conclusions and Future Work}
\label{sec:conclusion}
We focus on the consumer side problem under the realistic competitive market setting to design optimal bidding strategies (i.e., the optimal bidding functions).
This paper proposes Fed-Bidder to help DCs automatically determine their optimal bid prices for DOs in competitive market settings, taking into account the limited budget of DCs, eligible DOs and historical information (e.g., the prior distribution of the auctioned DO and the probability of winning the auction).
Fed-Bidder is, to the best of our knowledge, the first such approach which can support multiple DCs to simultaneously compete to attract DOs to join their respective federated learning effort. This enables auction-based FL client selection to operate in more realistic FL marketplaces.

In subsequent research, we will focus on extending Fed-Bidder to support more realistic scenarios in which each DO can simultaneously serve multiple FL model training tasks by multiplexing its local resources.

\section{Acknowledgments}
This research/project is supported by the National Research Foundation Singapore and DSO National Laboratories under the AI Singapore Programme (AISG Award No: AISG2-RP-2020-019); the RIE 2020 Advanced Manufacturing and Engineering (AME) Programmatic Fund (No. A20G8b0102), Singapore; the Nanyang Assistant Professorship (NAP); and Future Communications Research \& Development Programme (FCP-NTU-RG-2021-014).

\bibliographystyle{IEEEbib}
\bibliography{icme23}


\end{document}